\documentclass[11pt,a4paper]{article}

\usepackage[utf8]{inputenc}
\usepackage[T1]{fontenc}
\usepackage{lmodern}
\usepackage{geometry}
\usepackage{amsmath,amssymb}
\usepackage{graphicx}
\usepackage{subcaption}
\usepackage{booktabs}

\usepackage[numbers]{natbib}
\usepackage[]{hyperref}

\title{Comparing Post-Hoc Explainable AI Methods for Interpreting Black-Box EEG Models in Depression Detection}

\author{
    Antonia Šarčević, Nikolina Frid 
    \\
    University of Zagreb Faculty of Electrical Engineering and Computing, Croatia\\
    \texttt{antonia.sarcevic@fer.unizg.hr nikolina.frid@fer.unizg.hr}
}

\date{}

\begin{document}

\maketitle

\begin{abstract}
Recent advances in deep learning have enabled increasingly accurate electroencephalography (EEG)-based classification of Major Depressive Disorder (MDD), but the internal decision-making processes of high-capacity models remain difficult to interpret. In this study, we investigate the behavior of multiple post-hoc explainability methods applied to an InceptionTime architecture trained for EEG-based MDD detection. The analysis includes Shapley-based, gradient-based, and perturbation-based attribution approaches, namely DeepSHAP, Integrated Gradients, GradCAM, Occlusion, and Permutation Feature Importance.
Explainability analysis was performed within a subject-level stratified 5-fold cross-validation framework using global attribution aggregation across EEG segments and subjects. The evaluated methods revealed partially convergent attribution patterns, with recurring emphasis on frontal, temporal, and posterior EEG regions, particularly over the right hemisphere. Quantitative comparison further demonstrated substantial agreement between gradient- and perturbation-based approaches, while DeepSHAP produced comparatively distinct attribution distributions. At the same time, noticeable variability between explainability methods was observed, highlighting the influence of methodological assumptions on the resulting explanations.
Overall, the results suggest that different post-hoc explainability approaches may capture partially overlapping relevance structures when applied to EEG-based deep learning models for depression detection. Although the observed attribution patterns are broadly consistent with several findings previously reported in EEG studies of MDD, the present analysis should be interpreted as an exploratory investigation rather than evidence of definitive neurophysiological biomarkers or clinical applicability. The study highlights both the potential usefulness and the methodological limitations of post-hoc explainability for interpreting black-box EEG classifiers in psychiatric applications.
\end{abstract}


\section{Introduction}
Major depressive disorder (MDD) is among the most prevalent and disabling psychiatric disorders worldwide \citep{GBD2017IncidencePrevalence2018}. Diagnosis currently relies largely on symptom-based criteria, clinical interviews, and standardized rating scales, yet these approaches face longstanding concerns regarding diagnostic overlap, imperfect validity, and the difficulty of capturing clinically meaningful differences among patients \citep{Kendler2016}. These limitations are further compounded by symptom heterogeneity, overlap with other psychiatric conditions, variability in clinical presentation, and the absence of objective laboratory-based diagnostic markers, making MDD particularly challenging to diagnose and manage \citep{ZIMMERMAN201529,Rush2007,KesslerBromet2013}. Together, these challenges have motivated continued interest in whether neurophysiological measurements may provide complementary support for clinical decision-making.

Electroencephalography (EEG), as a non-invasive and widely accessible neuroimaging modality, has attracted sustained interest as a potential source of complementary information for MDD diagnosis and treatment-related assessment \citep{Kessler2018PredictiveAnalytics}. EEG has been investigated for diagnostic support, subtype characterization, and assessment of treatment-related changes. Numerous studies have explored spectral, connectivity, and nonlinear signal characteristics, with recurring reports of alterations in alpha-band measures, frontal alpha asymmetry, and theta- and gamma-related activity \citep{Kaiser31122018,Cukic2020EEGMDD,Simmatis2023EEGbiomarkersMDD}. However, no definitive EEG biomarker consensus has emerged, reflecting both the complexity of depression and substantial methodological variability across studies.

The possibility that clinically relevant information may reside not in a single biomarker, but in recurring signal patterns and potentially complex interactions among signal characteristics, has motivated increasing use of machine learning (ML) approaches for EEG analysis. Rather than relying solely on conventional analysis or isolated handcrafted descriptors, ML methods can exploit subtle multivariate patterns in EEG signals that may be difficult to characterize explicitly or detect using simpler analytical approaches \citep{Aleem2022,DEAGUIARNETO201983}. Both classical feature-based approaches \citep{HOSSEINIFARD2013339,MUMTAZ2017108,Cai2018} and more recent deep learning and time-series classification models have reported promising results in MDD detection \citep{avots2022ensemble,wang2024gctnet,dutua2025optimizing}. 

Recent high-capacity models have achieved promising performance in EEG-based MDD detection, but often at the cost of reduced interpretability. Many deep learning models used in biomedical time-series analysis rely on complex internal representations whose decision processes are not directly transparent \citep{JOVIC2025108153}. Concerns regarding the deployment of opaque models in high-stakes domains such as medicine are well established \citep{Rudin2019}, particularly in settings where understanding the basis of a prediction may be clinically relevant alongside predictive performance itself.

In this context, post-hoc explainability methods have emerged as an important approach for analyzing the behavior of black-box models \citep{Marcin2023,Linardatos2021}. Beyond improving transparency, explainability may also help relate model predictions to physiologically meaningful EEG characteristics and support broader clinical interpretation \citep{Hicks2021,Mulc2025}. However, different explainability methods rely on different assumptions and may emphasize different aspects of model behavior, potentially leading to divergent explanations \citep{JOVIC2025108153}. While previous studies have often focused on predictive performance or individual explainability techniques, fewer works have systematically examined the agreement and disagreement between multiple post-hoc explainability methods applied to the same EEG classification task.

In this work, we apply multiple post-hoc explainability methods to the InceptionTime model \citep{ismail2020inceptiontime} for EEG-based MDD detection. The goal is not to propose a new explainability approach, but to investigate what insights established post-hoc methods can provide when applied to a high-performing black-box time-series classification model. Specifically, we examine whether different explainability methods produce convergent or method-dependent patterns, and whether the resulting explanations support neurophysiologically plausible interpretations of model behavior. Through this analysis, the study aims to assess both the potential and the limitations of post-hoc explainability in the interpretation of black-box EEG models for depression detection.

The remainder of the paper is organized as follows. Section 2 provides background information on EEG-based depression detection, major categories of post-hoc explainability methods, and the use of explainability techniques in EEG studies. Section 3 describes the materials and methods used in the study, including the dataset, model architecture, and explainability framework. Section 4 presents the classification and explainability results. Section 5 discusses the obtained findings, limitations, and broader implications of the study. Finally, Section 6 concludes the paper and outlines potential directions for future work.

\section{Background}
\subsection{EEG-based detection of depression}
Interest in EEG-based depression detection has been driven by longstanding limitations of symptom-based psychiatric diagnosis, including diagnostic overlap, heterogeneity of clinical presentation, and the absence of objective laboratory-based markers \citep{Kendler2016,ZIMMERMAN201529,Rush2007}. As a non-invasive and widely accessible neuroimaging modality, EEG has therefore been investigated as a potential source of complementary information for diagnostic support and treatment-related assessment \citep{Kessler2018PredictiveAnalytics}. Numerous studies have reported recurring alterations in spectral, connectivity, and nonlinear signal characteristics, including alpha-band abnormalities, frontal alpha asymmetry, theta- and gamma-related activity, altered signal complexity, and disrupted large-scale coordination \citep{Kaiser31122018,Cukic2020EEGMDD,Simmatis2023EEGbiomarkersMDD,DeLaTorreLuque2017EEGMDD,Farzan2017EEGMDD,Chao2022EEGMDD}. Despite recurring findings, however, no definitive EEG biomarker consensus for MDD has emerged.

Most early EEG-based MDD detection approaches followed a classical machine learning pipeline consisting of feature extraction, feature selection, and classification. These methods typically relied on handcrafted spectral, statistical, nonlinear, or connectivity descriptors derived from previously reported EEG biomarkers or biomarker-like patterns \citep{cai2018study,wan2019single,sun2020study}. Conventional classifiers such as decision trees, logistic regression, K-nearest neighbors, and support vector machines offered at least partial interpretability through explicit features or simpler decision structures. Such approaches also enabled empirical comparison of candidate biomarkers across MDD subtypes or related psychiatric disorders, including bipolar disorder. However, reported performance was often highly dependent on feature design, dataset size, preprocessing choices, and evaluation protocols, limiting generalizability across studies.

More recent EEG studies increasingly employ automatic methods that treat neural signals as multivariate time series. Unlike classical pipelines based on handcrafted feature extraction, modern time-series classification (TSC) models can perform automatic feature extraction directly from raw signals \citep{burchert2024eeg}. By training on raw EEG recordings, such models can learn complex representations capturing multiscale temporal patterns, cross-channel interactions, and nonlinear characteristics that traditional handcrafted descriptors may overlook \citep{rafiei2022automated}. Recent studies have therefore increasingly employed higher-capacity models, including ensemble methods, graph-based neural architectures, convolutional neural networks, transformers, and time-series classification approaches operating on engineered or raw EEG representations \citep{movahed2021major,avots2022ensemble,wang2024gctnet,dutua2025optimizing,Liu2024}. These approaches can model complex multivariate interactions and temporal dependencies that may be difficult to capture using manually engineered descriptors alone, and have frequently reported stronger predictive performance. However, this increased predictive power has generally been accompanied by reduced interpretability, as the internal decision processes of deep or high-capacity models do not directly correspond to explicitly defined neurophysiological descriptors.

\subsection{Explainability in EEG studies}
The increasing use of high-capacity models for EEG-based MDD detection has also increased interest in explainability methods capable of probing otherwise opaque decision processes. In contrast to classical feature-based pipelines, the internal representations learned by deep neural networks and other high-capacity architectures do not directly correspond to explicitly defined neurophysiological descriptors, making interpretation substantially more difficult. As emphasized in \citep{JOVIC2025108153}, many high-performing models used in biomedical time-series analysis rely on complex transformations whose decision processes are not directly transparent. Moreover, as argued in \citep{Elton2020SAI}, deep neural networks operate through high-dimensional interpolation rather than stable human-interpretable rules, raising broader questions regarding the reliability and meaning of post-hoc explanations. Concerns surrounding the deployment of opaque models in high-stakes settings such as medicine are therefore well established \citep{Rudin2019,BurkartHuber2021,Rajpurkar2022}.

As a result, interpretation increasingly relies on post-hoc explainability techniques such as SHAP, Integrated Gradients, Grad-CAM, permutation importance, and occlusion-based analysis. Unlike intrinsically interpretable approaches, these explanations are inferred retrospectively through auxiliary analytical procedures rather than arising directly from the model structure itself. Beyond improving transparency, explainability in biomedical signal analysis may also support scientific insight by relating model predictions to physiologically meaningful EEG characteristics \citep{Hicks2021}. This is particularly relevant in psychiatric disorders such as MDD, where binary predictions alone may provide limited clinical value compared to identifying characteristic signal patterns, borderline cases, or features warranting further examination \citep{Mulc2025,Duran329}. Consequently, post-hoc explainability methods have become an increasingly important approach for analyzing black-box EEG models \citep{Marcin2023,Linardatos2021}.

Post-hoc explainability methods used in biomedical deep learning generally fall into several conceptual categories, including Shapley-value-based attribution, gradient-based attribution, and perturbation-based sensitivity analysis. Shapley-based approaches estimate feature contributions using concepts derived from cooperative game theory and provide theoretically grounded feature attributions \citep{lundberg2017unified}. Gradient-based methods analyze the sensitivity of model outputs with respect to changes in the input and are particularly well suited for differentiable architectures \citep{ancona2017towards,sundararajan2017axiomatic}. In contrast, perturbation-based approaches estimate feature importance by modifying parts of the input signal and observing the resulting changes in model predictions \citep{zeiler2014visualizing,ivanovs2021perturbation}. Although these approaches often provide complementary perspectives on model behavior, they also rely on substantially different assumptions and approximation strategies.

Most explainability approaches primarily generate local explanations describing the contribution of input features to individual predictions. However, broader understanding of model behavior often requires global explanations obtained by aggregating local explanations across multiple samples \citep{molnar2020interpretable}. In biomedical applications such as EEG-based MDD detection, such global explanations may help identify consistent signal patterns that align with existing neurophysiological knowledge.

At the same time, post-hoc explanations are not without limitations. Because these methods approximate model behavior rather than directly revealing internal decision-making processes, there is no guarantee that the generated explanations faithfully reflect the true reasoning underlying a prediction. Different explainability methods may therefore produce conflicting interpretations for the same sample, raising concerns regarding explanation stability, robustness, and physiological plausibility \citep{krishna2022disagreement,JOVIC2025108153}. Existing studies also commonly rely on a single explainability technique, leaving open the question of whether different post-hoc methods produce consistent explanations when applied to the same EEG classification problem.

Interpretation and comparison of reported EEG-based MDD detection results are further complicated by substantial variability in datasets and recording protocols. EEG recordings are highly sensitive to acquisition conditions, recording hardware, preprocessing pipelines, subject state, medication effects, and experimental design. Many studies also rely on relatively small private cohorts, often involving between 15 and 60 subjects \citep{Mahato2019,Liu2022,Zhu2020,avots2022ensemble}, while only a limited number of public datasets, such as \citep{Mumtaz2016}, are widely reused. Combining datasets across studies is often difficult due to differences in electrode configurations, recording procedures, diagnostic criteria, preprocessing strategies, and population characteristics, making direct comparison of reported performance metrics problematic. Additional methodological concerns arise from evaluation procedures themselves, particularly in studies where recordings are segmented into multiple samples, as inadequate separation of subject-wise and sample-wise splits may introduce subject leakage and artificially inflate reported performance.

\section{Materials and Methods}

\subsection{Dataset description}
To evaluate explainable artificial intelligence (XAI) methods in the context of EEG-based depression detection, we utilized a dataset originally described by Mulc et al.~\cite{Mulc2025}. Data acquisition was conducted at the University Psychiatric Hospital Vrapče in accordance with the Declaration of Helsinki and with approval from the hospital’s ethics committee.

The dataset consists of electroencephalogram (EEG) recordings from 140 adult participants ($\ge 18$ years), evenly divided into a patient group and a healthy control group. The patient group includes 70 individuals diagnosed with moderate-to-severe Major Depressive Disorder (MDD) according to ICD-10 criteria. Participants with additional psychiatric diagnoses (except personality disorders), neurological disorders, or major somatic conditions were excluded. Mild and controlled conditions such as hypertension and hyperlipidemia were permitted. The control group consists of 70 healthy participants without psychiatric or neurological disorders and without psychopharmacological treatment. The groups were approximately matched by age and sex. The MDD group includes 37 women with a mean age of $36.86 \pm 10.22$ years and 33 men with a mean age of $45.24 \pm 12.10$ years, while the healthy control group includes 32 women aged $35.88 \pm 9.96$ years and 38 men aged $36.16 \pm 11.35$ years.

\begin{figure}[h]
    \centering

    \begin{subfigure}[]{0.35\textwidth}
        \centering
        \includegraphics[width=\textwidth]{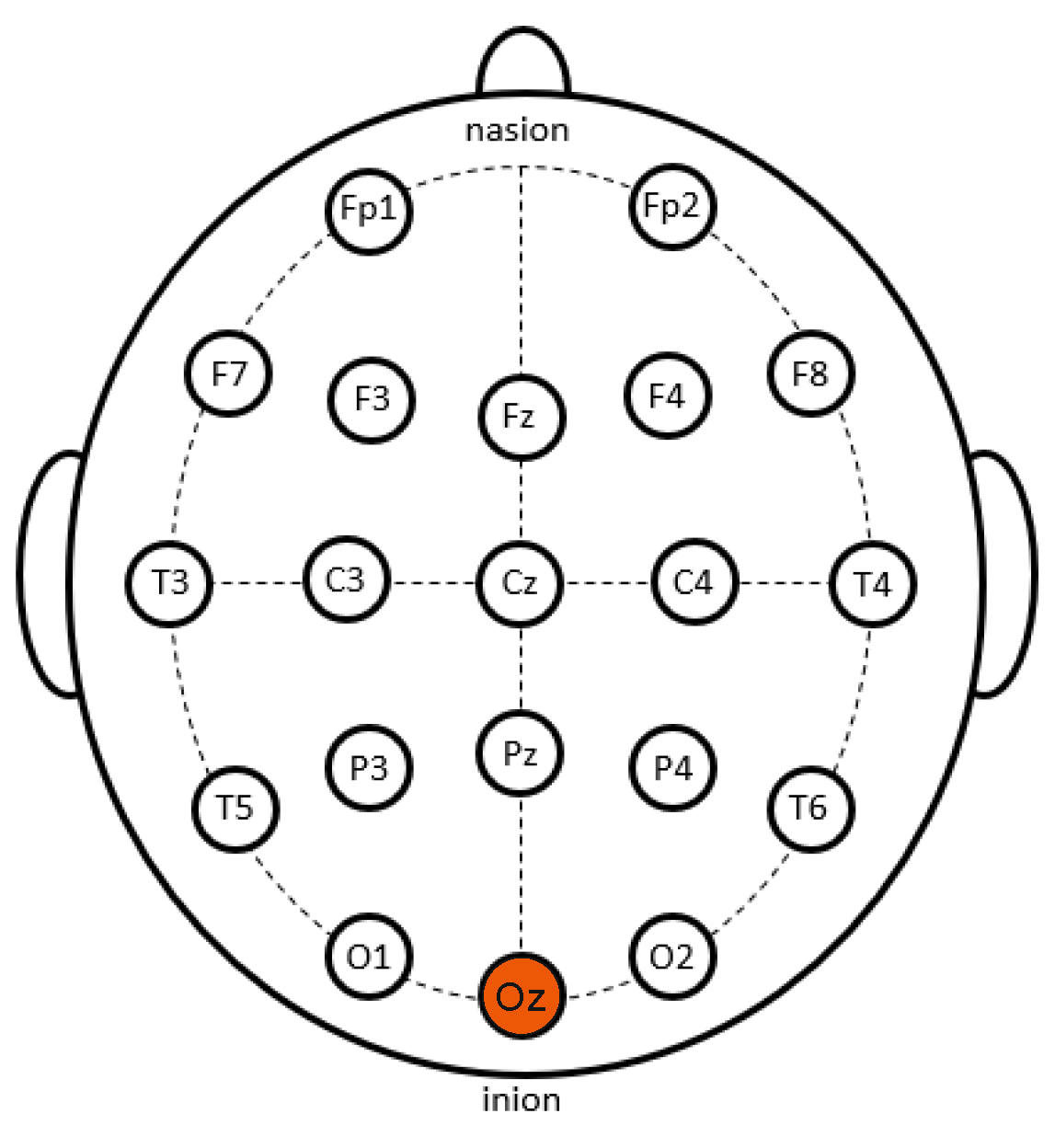}
        \caption{}
        \label{fig:eeg10-20}
    \end{subfigure}%
    \begin{subfigure}[]{0.4\textwidth}
        \centering
        \includegraphics[width=\textwidth]{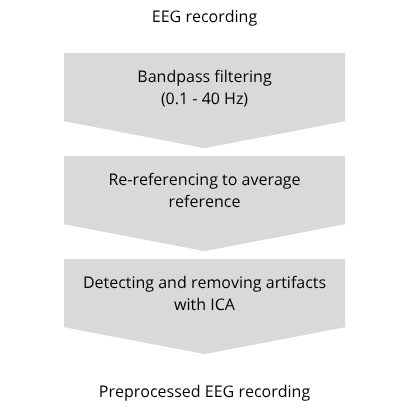}
        \caption{}
        \label{fig:eeg_preprocessing}
    \end{subfigure}

    \caption{\textbf{a}) EEG 10–20 system with 19 channels 
    (\textbf{b}) EEG preprocessing protocol}
    \label{fig:eeg-placement}
\end{figure}

All recordings were acquired using a Nihon Kohden EEG-1200K clinical device at a sampling frequency of 200~Hz. Scalp potentials were recorded from 19 channels arranged according to the standard 10--20 system with an Oz reference electrode (Figure~\ref{fig:eeg-placement}). Each participant underwent a 30-minute recording session in a supine position in a quiet, dimly lit environment. The protocol included alternating eyes-open and eyes-closed resting states, photo-stimulation at frequencies of 4, 8, 16, 24, and 30~Hz, and a period of induced hyperventilation. Relevant events such as eye movements, blinks, and muscle activity were annotated during acquisition to facilitate artifact identification.

Because EEG recordings are frequently contaminated by physiological and environmental artifacts, preprocessing was performed following the protocol described in~\cite{Mulc2025}. The preprocessing was conducted in MATLAB R2021b using the EEGLAB toolbox. Signals were first filtered using a finite-impulse-response (FIR) band-pass filter between 0.1 and 40~Hz, followed by common-average re-referencing. Independent component analysis (ICA) was then applied to decompose the signal, after which ICLabel was used to automatically classify components into categories such as brain, eye, muscle, heart, and line noise. Components identified with high confidence as artifacts were removed while preserving neural activity-related information.

\subsection{Model architecture}
\begin{figure}[!h]
    \centering
    \includegraphics[width=1\textwidth]{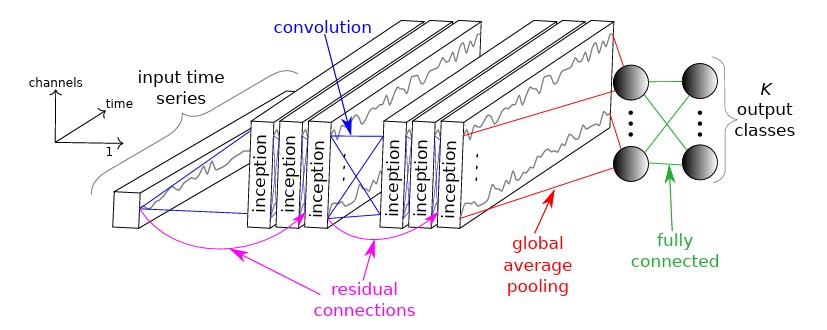}
    \caption{InceptionTime architecture for time-series classification adapted from \cite{ismail2020inceptiontime}.}
    \label{fig:inception-time}
\end{figure}

To evaluate post-hoc explainability methods for EEG-based depression classification, the InceptionTime architecture \citep{ismail2020inceptiontime} was selected due to its strong performance and established robustness in multivariate time-series classification tasks, including recent EEG applications. InceptionTime is a deep convolutional architecture designed for automatic time-series classification. The model utilizes Inception modules containing parallel convolutions with different filter lengths, enabling extraction of temporal patterns at multiple scales and allowing the network to capture discriminative signal characteristics of varying duration and frequency. Residual connections are employed to facilitate optimization and improve training robustness, while final predictions are obtained through global average pooling followed by a classification layer. A schematic overview of the architecture is shown in Figure~\ref{fig:inception-time}.

Input EEG recordings were restricted to the first 300 seconds and segmented into overlapping 2-second windows with 0.5 seconds overlap between consecutive segments. Given the acquisition frequency of 200~Hz, each epoch contained 400 temporal samples per channel, resulting in input tensors of dimensions $(19 \times 400)$. Overlapping segmentation increased the number of available training samples while preserving temporal continuity of the EEG signal. The EEG recordings were treated as multivariate time series, where each channel represented one input feature dimension and temporal convolutions were applied jointly across channels. Prior to training, channel-wise z-score normalization parameters were computed exclusively from the training data within each fold and subsequently applied to validation and test sets to avoid information leakage. This normalization procedure reduced inter-subject variability in signal amplitude and improved training stability by ensuring comparable feature scales across both subjects and channels.

Model evaluation was performed using stratified 5-fold cross-validation at the subject level, ensuring that all EEG segments belonging to a single participant remained within the same split. Stratification preserved class balance between healthy controls and MDD participants across folds, while preventing subject leakage between training and testing data, which is particularly important in EEG analysis due to strong intra-subject similarity between neighboring signal segments. Preventing such leakage is critical because sample-wise splitting may artificially inflate classification performance in EEG-based studies. A separate model was independently trained within each fold, resulting in a total of five trained models across the complete cross-validation procedure. Within each training fold, 10\% of subjects were further reserved for validation and model selection.

The model was implemented using the standard configuration proposed in \citep{ismail2020inceptiontime}, consisting of six inception blocks ($\text{depth}=6$) and 64 convolutional filters per block ($\text{nf}=64$). Bottleneck layers were enabled to reduce the number of trainable parameters. Training was performed using the Adam optimizer with cross-entropy loss and the One-Cycle learning-rate scheduling policy. The maximum learning rate was set to $3 \times 10^{-3}$ with a batch size of 64 over 10 training epochs. Training was conducted for a fixed number of epochs, after which the model state associated with the lowest validation loss was retained for subsequent explainability analysis.

\subsection{Explainability framework}
To analyze the decision-making behavior of the trained InceptionTime models, a post-hoc explainability framework combining Shapley-value-based, gradient-based, and perturbation-based attribution methods was implemented. Since EEG-based deep learning models operate on high-dimensional spatiotemporal signals in which informative patterns may be distributed across multiple channels and temporal intervals, direct interpretation of learned representations is difficult without post-hoc analysis. The framework was therefore designed to compare explanations produced under different interpretability assumptions while evaluating the stability of the resulting attribution patterns across cross-validation folds.

For a given EEG segment, the input sequence is represented as $X \in \mathbb{R}^{C \times T}$, where $C = 19$ denotes the number of EEG channels and $T = 400$ represents the temporal dimension. Each explainability method produces a local attribution representation associated with the input signal, indicating the relative contribution of individual channel--time regions to the model prediction. To obtain global explanations, absolute attribution values were first averaged across temporal samples within each EEG segment and subsequently aggregated at the subject level. The resulting subject-level attribution profiles were then averaged across cross-validation folds to produce fold-aggregated channel-level importance estimates.

\subsubsection{DeepSHAP}

DeepSHAP explanations were generated using Captum's \texttt{DeepLiftShap} implementation integrated with the PyTorch-based InceptionTime model \citep{lundberg2017unified,shrikumar2017learning}. DeepSHAP combines Shapley-value approximation with DeepLIFT-style backpropagation rules to efficiently estimate feature contributions in deep neural networks. The method estimates feature relevance relative to a reference baseline dataset by propagating contribution scores through the network.

To approximate baseline expectations while avoiding excessive computational overhead and potential out-of-distribution artifacts, a background reference set consisting of $K = 40$ randomly selected training samples was generated independently within each cross-validation fold. Smaller background sets may produce unstable Shapley estimates due to noisy baseline expectations, whereas using the full training partition would substantially increase computational cost. Attribution maps were computed for $M = 4000$ test samples across all folds. Absolute attribution values were first averaged across temporal samples within each EEG segment, subsequently aggregated at the subject level, and finally averaged across folds to obtain global channel-importance rankings. Although DeepSHAP provides computationally efficient approximations of Shapley-value-based explanations, the resulting attribution profiles may remain sensitive to the selection of the reference baseline dataset.

\subsubsection{Integrated Gradients}
Integrated Gradients estimates feature relevance by integrating model gradients along a continuous interpolation path between a reference baseline and the input signal, thereby reducing some limitations of standard gradient-based attribution methods, such as gradient saturation and local noise sensitivity. We used Captum attribution framework \citep{sundararajan2017axiomatic} for computing Integrated Gradients explanations. 

The path integral approximation was evaluated using $S = 100$ interpolation steps between the baseline and the input signal. This configuration provided a practical trade-off between numerical approximation accuracy and computational efficiency. Attribution maps were computed for $M = 4000$ true-positive test samples within the cross-validation framework.

To examine the influence of baseline selection, two reference configurations were evaluated: a zero-valued baseline and a Gaussian-noise baseline generated using the empirical mean and variance of the training data within each fold. While the zero baseline represents complete signal absence, the Gaussian-noise baseline preserves the general statistical properties of EEG recordings while disrupting structured temporal patterns. Since Integrated Gradients explanations may remain sensitive to baseline selection, both configurations were included for comparative stability analysis. Absolute attribution values were first averaged across temporal samples within each EEG segment, subsequently aggregated at the subject level, and finally averaged across folds to obtain fold-aggregated channel-level importance estimates.

\subsubsection{GradCAM}
Grad-CAM analysis was applied to the output of the final Inception block within the InceptionTime architecture prior to global average pooling \citep{selvaraju2017grad}. This layer represents the highest level of abstract temporal feature encoding before dimensionality reduction through pooling, while still preserving temporal structure within the learned representations. ReLU activation was used to retain features positively associated with the target class.

Because the internal feature representations operate at reduced temporal resolution, the resulting attribution maps were linearly interpolated back to the original input length of $T = 400$. Since Grad-CAM primarily provides coarse temporal localization and lacks precise electrode-level attribution, the analysis was additionally supplemented with InputXGradient attribution. InputXGradient computes feature relevance by multiplying the input signal with the corresponding input gradients ($X \times \nabla_X f(X)$), enabling estimation of channel-level contribution patterns directly in the input space. Combining Grad-CAM with InputXGradient therefore enables simultaneous analysis of high-level activation regions and finer input-level attribution patterns.

\subsubsection{Occlusion analysis}
Occlusion-based sensitivity analysis was implemented using a sliding-window perturbation approach operating without gradient backpropagation \citep{zeiler2014visualizing}. Local signal regions were replaced with zero-valued masks and the resulting reduction in target-class probability was measured. Occlusion windows of size $(1,50)$ with a stride of $(1,25)$ were applied across the temporal dimension of each EEG channel, producing a 50\% overlap between neighboring perturbation regions. The overlapping configuration improved localization continuity and reduced abrupt transitions between adjacent occlusion windows.

The selected window size provided a compromise between temporal localization and preservation of informative signal structure, since excessively large windows may obscure localized relevance patterns while excessively small windows may fail to sufficiently perturb distributed temporal features. Absolute sensitivity values were first aggregated across temporal regions within each EEG segment, subsequently aggregated at the subject level, and finally averaged across folds to obtain global channel-level importance estimates.

\subsubsection{Permutation feature importance}
Permutation feature importance (PFI) was implemented using Captum's \texttt{FeaturePermutation} module \citep{breiman2001random,fisher2019all}. Individual EEG channels were independently permuted across the evaluation batch in order to disrupt temporal structure while preserving marginal value distributions. Unlike zero-masking approaches, permutation-based perturbation preserves the overall statistical properties of the signal while removing temporally structured information associated with the selected channel.

Channel importance was quantified using the absolute decrease in target-class probability after permutation. Each channel was evaluated independently in order to estimate its relative contribution to the model prediction under temporally corrupted input conditions. Absolute attribution values were subsequently aggregated at the subject level and averaged across folds to obtain global channel-level importance estimates.

\subsubsection{Stability and reliability analysis}

To evaluate the robustness and consistency of the generated explanations, multiple complementary stability and agreement metrics were computed. The evaluation framework considers three main aspects of explainability behavior: fold-wise stability, attribution compactness, and cross-method ranking agreement.

Fold-wise stability was quantified using the coefficient of variation ($CV$), where lower values indicate more stable attribution patterns across cross-validation splits. Attribution compactness was evaluated using the Gini sparsity coefficient ($G$), which measures whether attribution values are diffusely distributed or concentrated within a smaller subset of highly relevant channels. Agreement between the highest-ranked feature subsets produced by different explainability methods was assessed using the Jaccard index across top-$K$ channel rankings ($K \in \{5,10,15\}$), while global ranking consistency across complete feature rankings was evaluated using Kendall's rank correlation coefficient ($\tau$).

Cross-method ranking consistency was further evaluated using Sequential Rank Agreement (SRA) \citep{SRAEkstrom2019}, which measures progressive agreement between ranked feature lists using overlap-based similarity across increasing top-$K$ subsets. In addition to channel-level SRA, a regional variant was computed after aggregating channel attributions into predefined cortical regions using mean regional importance values. In addition, a spatial-temporal variation (STV) metric adapted from graph-based total variation measures \citep{STVMortaheb} was used to evaluate smoothness and local variability within the generated attribution maps. The metric computes normalized L1 differences between neighboring EEG channels defined according to the standard 10--20 electrode topology, where lower STV values indicate smoother and spatially more coherent attribution distributions.

By combining complementary metrics capturing different aspects of attribution behavior, the framework enables assessment of both the stability of individual explainability methods and the consistency of explanations produced across different attribution approaches. This multi-metric evaluation reduces the likelihood that apparently stable attribution patterns arise from isolated methodological artifacts or metric-specific biases.

\section{Results}

\subsection{Classification Performance}
\begin{figure}[h]
    \centering

    \begin{subfigure}[b]{0.48\textwidth}
        \centering
        \includegraphics[width=\textwidth]{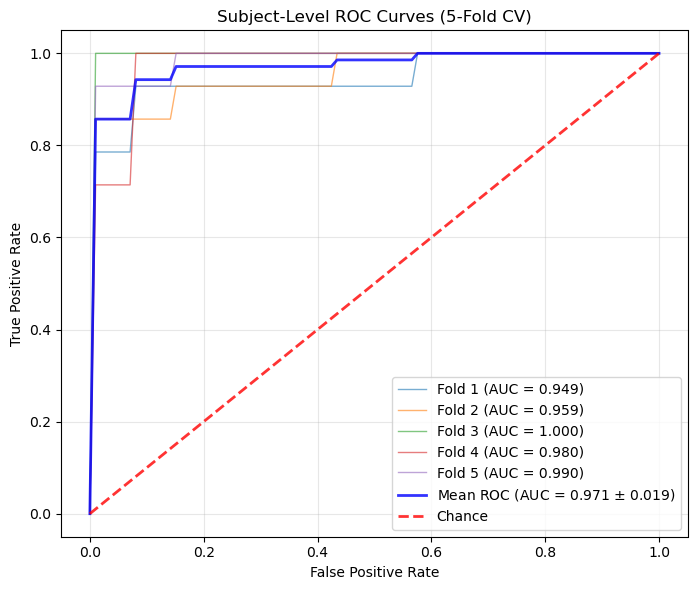}
        \caption{ROC curve for InceptionTime evaluation.}
        \label{fig:roc}
    \end{subfigure}
    \hfill
    \begin{subfigure}[b]{0.48\textwidth}
        \centering
        \includegraphics[width=\textwidth]{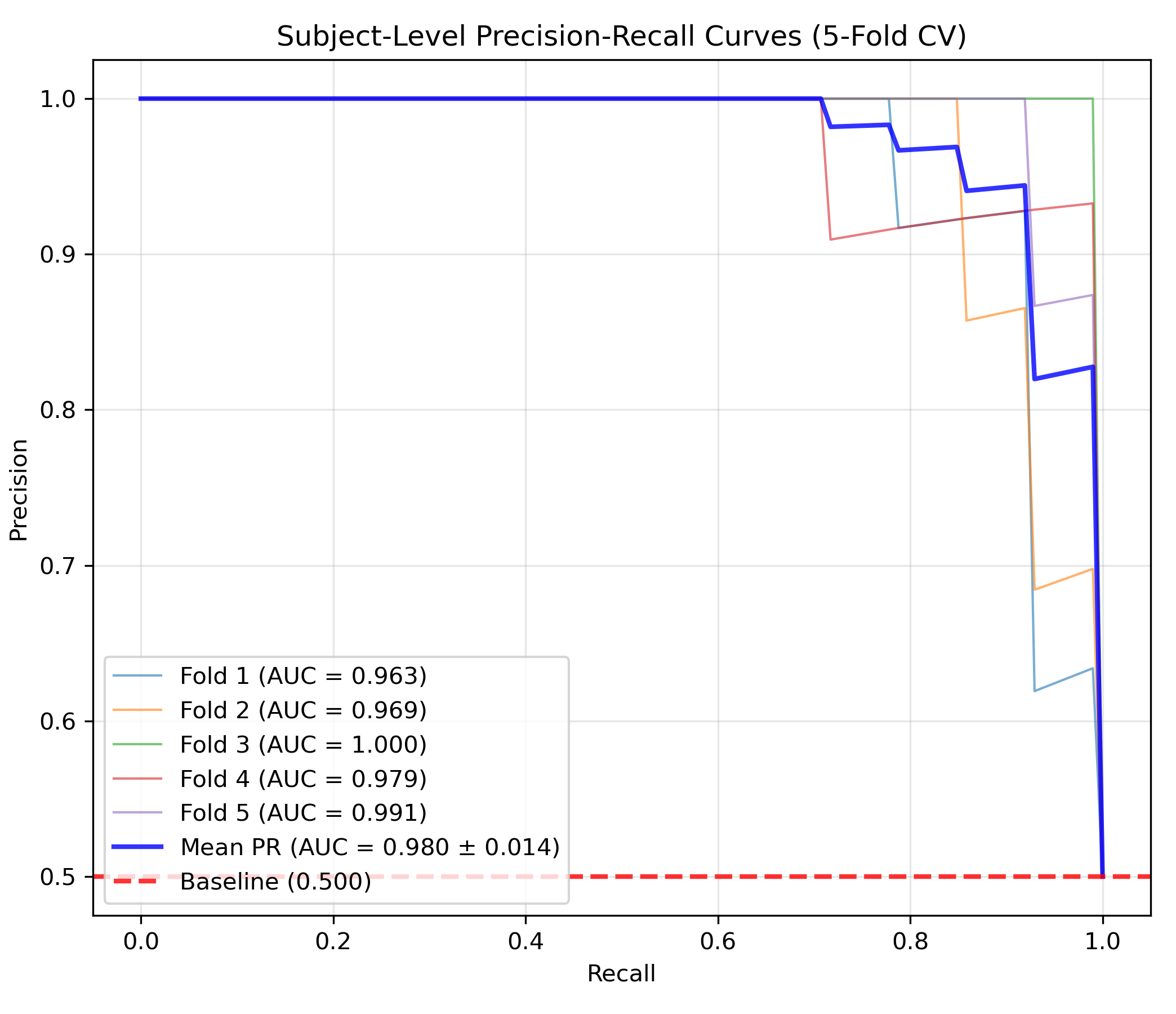}
        \caption{Precision--Recall curve for InceptionTime evaluation.}
        \label{fig:pr}
    \end{subfigure}

    \caption{Performance evaluation curves for the InceptionTime model.}
    \label{fig:both}
\end{figure}

The quantitative performance of the InceptionTime architecture was evaluated using subject-level aggregated metrics within a stratified 5-fold cross-validation framework. To obtain subject-level predictions, the probabilities generated for all epochs belonging to a participant were averaged, and the final class label was assigned according to the maximum value of the resulting probability vector. The complete performance metrics for each fold, together with the overall mean and standard deviation, are presented in Table~\ref{tab:inception_time_results}, while the corresponding subject-level ROC and PR curves are illustrated in Figures~\ref{fig:roc} and~\ref{fig:pr}, respectively.

Overall, the model achieved consistently strong classification performance across all folds, with balanced accuracy, precision, recall, and F1-score values. The ROC AUC and PR AUC results further indicate good discriminative capability between healthy controls and individuals with Major Depressive Disorder (MDD). Although a moderate reduction in performance was observed in Fold 4, the corresponding AUC values remained relatively high, suggesting that the model preserved stable class separation despite increased variability within that test partition.

The relatively low standard deviations across folds indicate reasonably consistent generalization performance under the adopted evaluation protocol. Since the classifier achieved stable predictive performance across multiple cross-validation splits, the resulting attribution maps provide a suitable basis for subsequent post-hoc explainability analysis. However, given the limited dataset size and the variability commonly observed in EEG-based depression studies, the reported performance should be interpreted as exploratory rather than definitive evidence of clinical applicability.

\begin{table}[ht]
\centering
\caption{Inception Time 5-Fold Cross-Validation Results}
\label{tab:inception_time_results}
\begin{tabular}{lccccccc}
\hline
\textbf{Metric} & \textbf{Fold 1} & \textbf{Fold 2} & \textbf{Fold 3} & \textbf{Fold 4} & \textbf{Fold 5} & \textbf{Mean} & \textbf{STD} \\
\hline
Accuracy  & 0.9286 & 0.9286 & 0.9643 & 0.8214 & 0.9286 & 0.9143 & 0.0484 \\
Precision & 0.9286 & 0.9375 & 0.9667 & 0.8684 & 0.9286 & 0.9259 & 0.0320 \\
Recall    & 0.9286 & 0.9286 & 0.9643 & 0.8214 & 0.9286 & 0.9143 & 0.0484 \\
F1-score  & 0.9286 & 0.9282 & 0.9642 & 0.8155 & 0.9286 & 0.9130 & 0.0507 \\
ROC AUC   & 0.9490 & 0.9592 & 1.0000 & 0.9796 & 0.9898 & 0.9755 & 0.0189 \\
PR AUC    & 0.9634 & 0.9690 & 1.0000 & 0.9787 & 0.9911 & 0.9804 & 0.0135 \\
\hline
\end{tabular}
\end{table}

\subsection{Explainability Results}

\subsubsection{Global attribution patterns}
For each explainability method, attribution maps were initially computed at the level of individual EEG segments and subsequently aggregated across subjects and cross-validation folds to obtain global class-level relevance profiles for the Control and MDD groups. Preliminary temporal attribution analysis was also performed in an attempt to identify localized transient patterns or event-specific relevance structures. However, the resulting temporal attributions were comparatively unstable and did not reveal consistently interpretable short-term patterns across folds or subjects. Consequently, the following analysis focuses primarily on aggregated spatial and spectral attribution trends, where the explainability methods exhibited substantially greater consistency and clearer separation between the two diagnostic groups.

\begin{figure}[!h]
    \centering
    \includegraphics[width=0.9\linewidth]{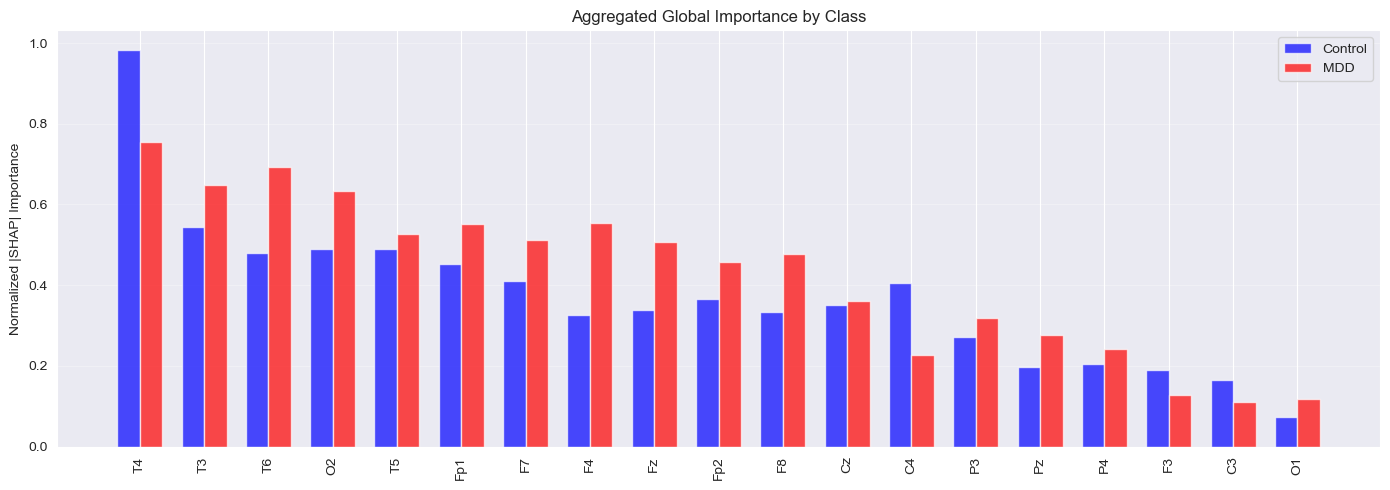}
    \caption{Global channel importance profile generated using DeepSHAP, comparing mean attribution weights across all EEG channels for the Control and Major Depressive Disorder (MDD) groups.}
    \label{fig:deep_channel}
\end{figure}

The global attribution analysis revealed clear differences in channel-level importance patterns between the Control and MDD groups across both evaluated explainability methods. DeepSHAP attribution profiles, shown in Figure~\ref{fig:deep_channel}, indicate that the right temporal electrode T4 exhibited high attribution values in both groups, while the MDD condition demonstrated broader attribution emphasis across right-hemisphere channels, particularly T6, O2, F4, and the midline frontal channel Fz.

\begin{figure}[!h]
    \centering
    \includegraphics[width=0.85\linewidth]{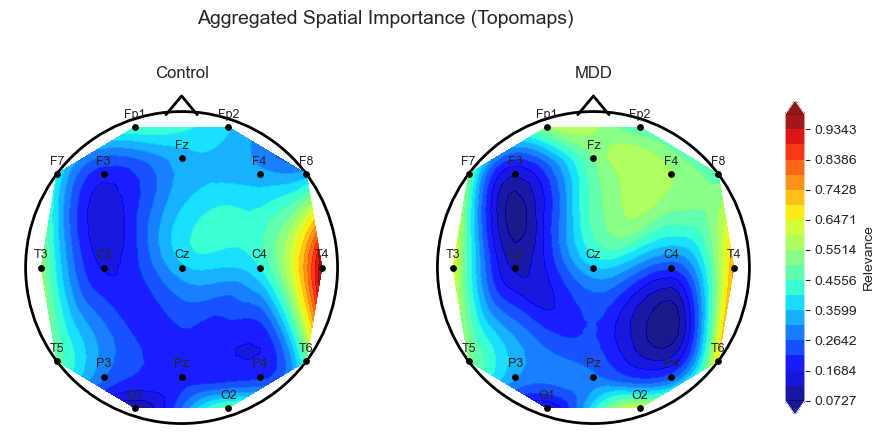}
    \caption{Topographic mapping of DeepSHAP feature attributions, illustrating the spatial distribution of model relevance across the scalp for both Control and MDD classes.}
    \label{fig:deep_topo}
\end{figure}

The corresponding topographic maps presented in Figure~\ref{fig:deep_topo} further illustrate these differences. In the Control condition, the attribution profile remains comparatively localized around the right temporal region, whereas the MDD condition exhibits a broader spatial distribution extending across frontal and posterior regions.

\begin{figure}[!h]
    \centering
    \includegraphics[width=0.9\linewidth]{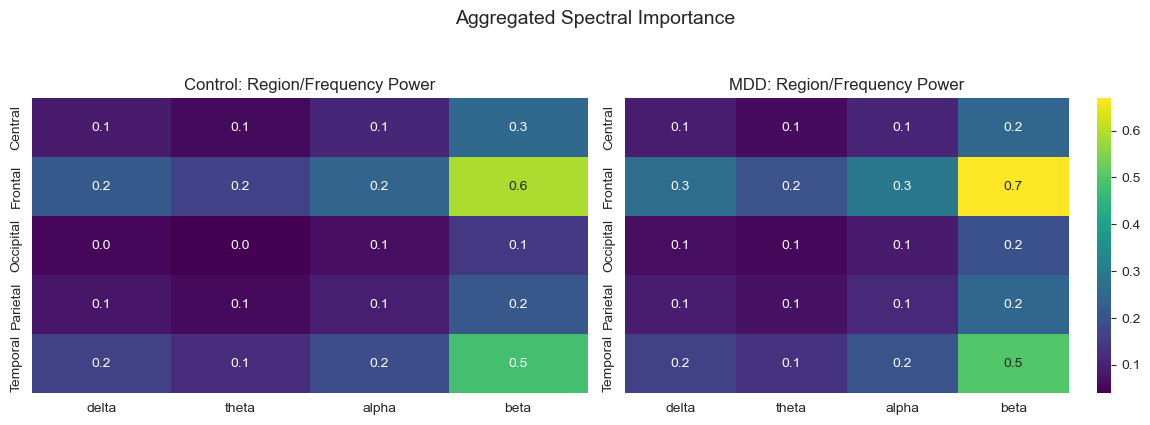}
    \caption{DeepSHAP regional and frequency-band heatmap matrix illustrating the interaction between cortical regions and EEG frequency bands (delta, theta, alpha, beta, gamma) in driving model predictions.}
    \label{fig:deep_heatmap}
\end{figure}

The regional-frequency attribution maps generated using DeepSHAP (Figure~\ref{fig:deep_heatmap}) indicate that beta-band activity contributed substantially to model predictions in both groups, while the MDD condition additionally showed elevated attribution values across slower delta-, theta-, and alpha-band components concentrated within frontal regions.

\begin{figure}[!h]
    \centering
    \includegraphics[width=0.9\linewidth]{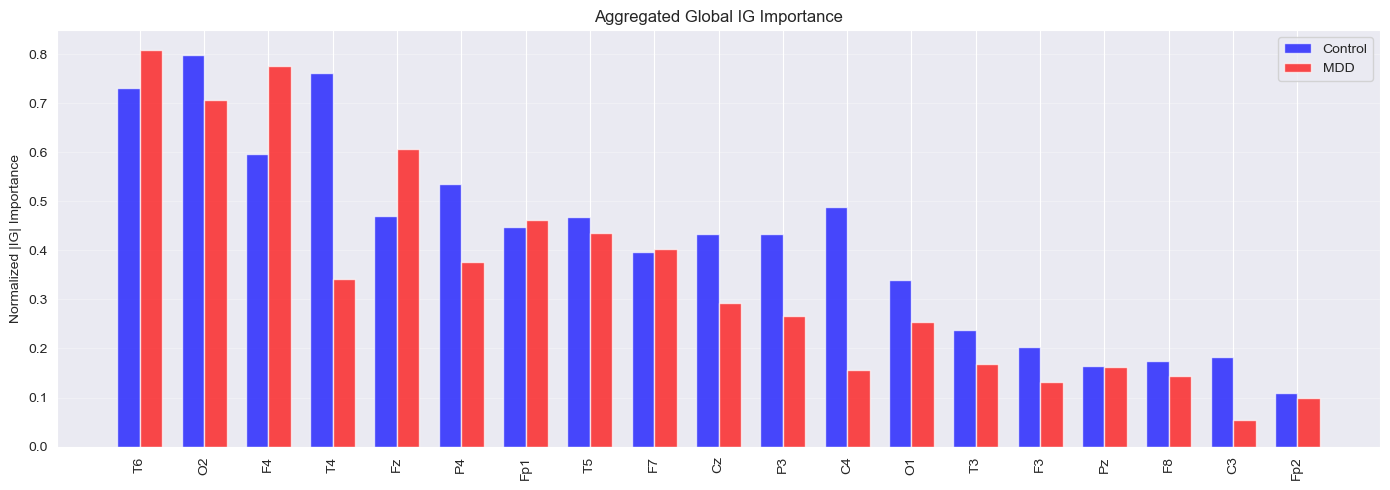}
    \caption{Global channel importance profile generated using Integrated Gradients (IG), comparing mean attribution scores across EEG channels for the Control and MDD groups.}
    \label{fig:ig_channel}
\end{figure}

Integrated Gradients (IG) produced attribution patterns broadly consistent with those observed using DeepSHAP while yielding somewhat stronger contrast between the two classes. As shown in Figure~\ref{fig:ig_channel}, T6, O2, F4, and T4 emerged as dominant channels, with increased attribution emphasis over right frontal and temporal-parietal regions in the MDD group.

\begin{figure}[!h]
    \centering
    \includegraphics[width=0.85\linewidth]{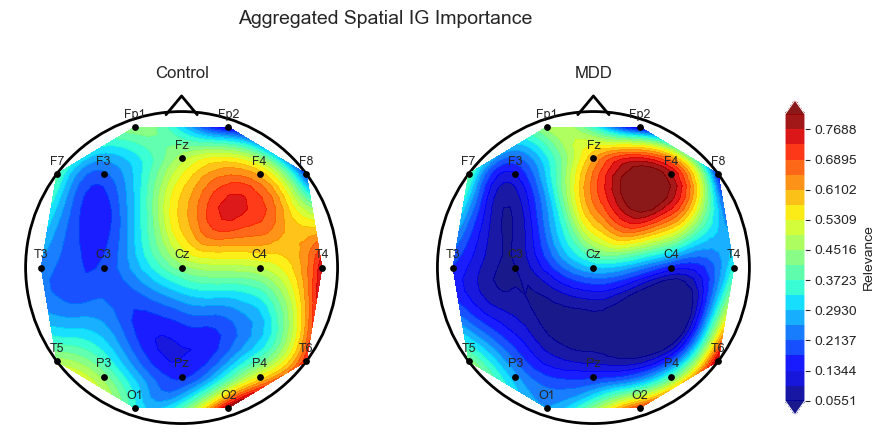}
    \caption{Topographic projection of Integrated Gradients (IG) feature attributions, illustrating distinct spatial relevance patterns between the Control and MDD groups.}
    \label{fig:ig_topo}
\end{figure}

The corresponding topographic projections shown in Figure~\ref{fig:ig_topo} similarly indicate concentrated attribution regions over right frontal and posterior areas.

\begin{figure}[!h]
    \centering
    \includegraphics[width=0.9\linewidth]{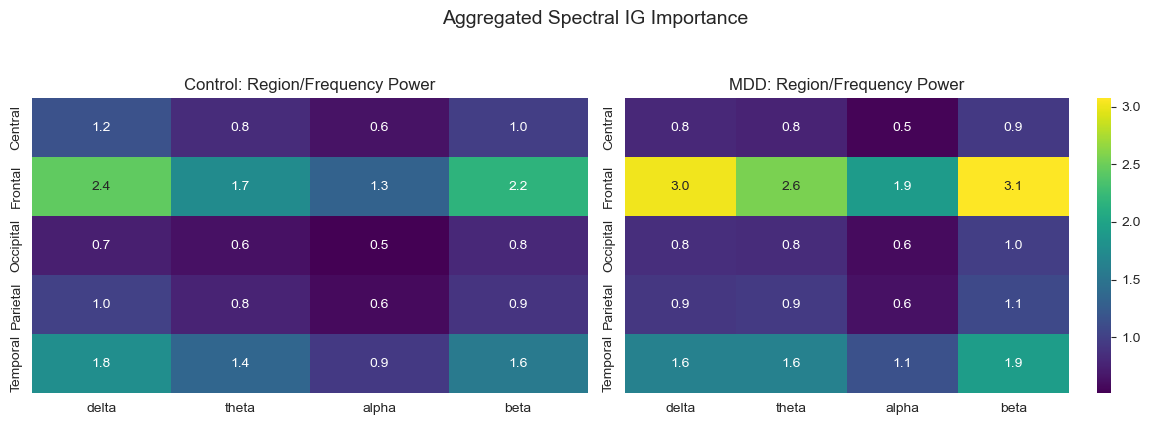}
    \caption{Integrated Gradients (IG) regional and frequency-band attribution heatmap evaluating the contribution of joint spatial-spectral features across both study groups.}
    \label{fig:ig_heatmap}
\end{figure}

The regional-frequency heatmaps generated using IG (Figure~\ref{fig:ig_heatmap}) further demonstrate that frontal and temporal regions contributed prominently across multiple frequency bands, with elevated delta- and beta-band attribution values observed in the MDD condition relative to Controls.

\begin{figure}[!h]
    \centering
    \includegraphics[width=0.9\linewidth]{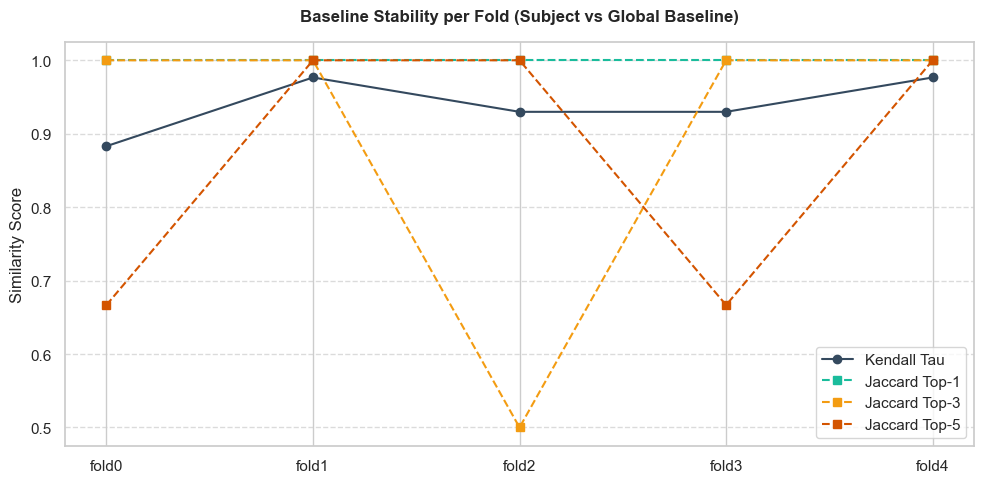}
    \caption{Integrated Gradients baseline stability analysis across the 5-fold cross-validation scheme, showing Kendall’s Tau and Jaccard similarity metrics for Top-1, Top-3, and Top-5 features.}
    \label{fig:ig_stability}
\end{figure}

To evaluate the sensitivity of IG explanations to baseline selection, an additional baseline stability analysis was performed using both subject-specific local baselines and a global zero-signal baseline. As illustrated in Figure~\ref{fig:ig_stability}, Kendall’s Tau values remained consistently high across folds, while Jaccard similarity scores for the highest-ranked channels showed relatively limited variation. These results suggest that the dominant attribution patterns remained reasonably stable under the evaluated baseline configurations.

\subsubsection{Stability and cross-method agreement}
\begin{figure}[h]
    \centering
    \includegraphics[width=0.75\textwidth]{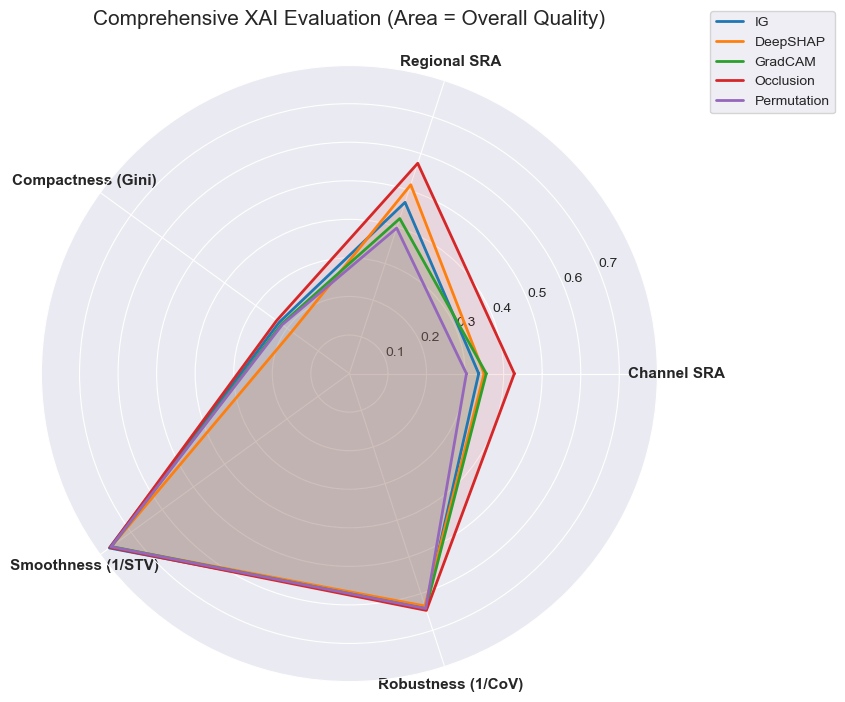}
    \caption{Multi-metric radar chart comparing Integrated Gradients, DeepSHAP, GradCAM, Occlusion, and Permutation across SRA, Regional SRA, CoV, Gini, and STV metrics.}
    \label{fig:radar}
\end{figure}

Quantitative stability and agreement metrics for the evaluated explainability methods are summarized in Table~\ref{tab:xai_stability} and visualized in Figure~\ref{fig:radar}.

\begin{table}[h]
\centering
\caption{Quantitative evaluation of explainability-method stability and agreement using Sequential Rank Agreement (SRA), regional SRA, coefficient of variation (CoV), Gini sparsity coefficient, and spatial-temporal variation (STV).}
\label{tab:xai_stability}
\begin{tabular}{lccccc}
\toprule
Method & SRA & Regional SRA & CoV & Gini & STV \\
\midrule
IG & 0.335 & 0.467 & 0.565 & 0.226 & 0.299 \\
DeepSHAP & 0.349 & 0.515 & 0.578 & 0.185 & 0.308 \\
GradCAM & 0.355 & 0.423 & 0.560 & 0.218 & 0.310 \\
Occlusion & 0.428 & 0.574 & 0.548 & 0.234 & 0.303 \\
Permutation & 0.303 & 0.397 & 0.561 & 0.216 & 0.308 \\
\bottomrule
\end{tabular}
\end{table}

Among the evaluated methods, Occlusion achieved the highest Sequential Rank Agreement (SRA = 0.428) and Regional SRA (0.574), indicating comparatively consistent ranking behavior under contiguous perturbation analysis. DeepSHAP produced the lowest Gini coefficient (0.185), corresponding to a more concentrated attribution distribution relative to the remaining methods.

The coefficient of variation (CoV) values remained relatively similar across all evaluated approaches, ranging between 0.548 and 0.578. Likewise, the Spatial-Temporal Variation (STV) metric exhibited only limited variation across methods. Overall, these results suggest that the explainability methods produced attribution structures of broadly comparable smoothness and fold-wise variability under the adopted evaluation framework.

\begin{figure}[!h]
    \centering
    \includegraphics[width=0.75\textwidth]{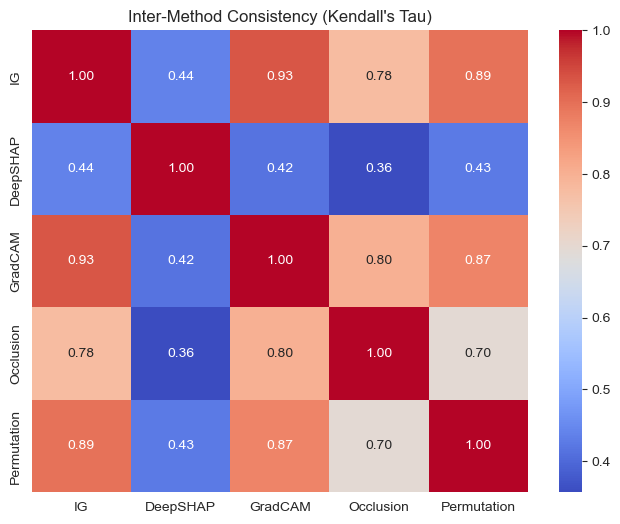}
    \caption{Inter-method consistency heatmap illustrating Kendall’s Tau cross-correlation values between the five evaluated explainability frameworks.}
    \label{fig:corr_matrix}
\end{figure}

Agreement between explainability methods was further evaluated using pairwise Kendall’s Tau correlations, illustrated in Figure~\ref{fig:corr_matrix}. The strongest agreement was observed between IG and GradCAM ($\tau = 0.93$), while IG and Permutation ($\tau = 0.89$) as well as GradCAM and Permutation ($\tau = 0.87$) also demonstrated substantial ranking similarity. In contrast, DeepSHAP exhibited noticeably lower agreement with the remaining approaches, with correlation values ranging from 0.36 to 0.44.

These results indicate that gradient-based and perturbation-based methods produced partially convergent attribution rankings, whereas the Shapley-based formulation of DeepSHAP yielded comparatively distinct feature attribution profiles. The observed differences highlight the influence of underlying methodological assumptions on the resulting explanations.

\subsubsection{Consensus attribution profile}
\begin{figure}[!h]
    \centering
    \includegraphics[width=\linewidth]{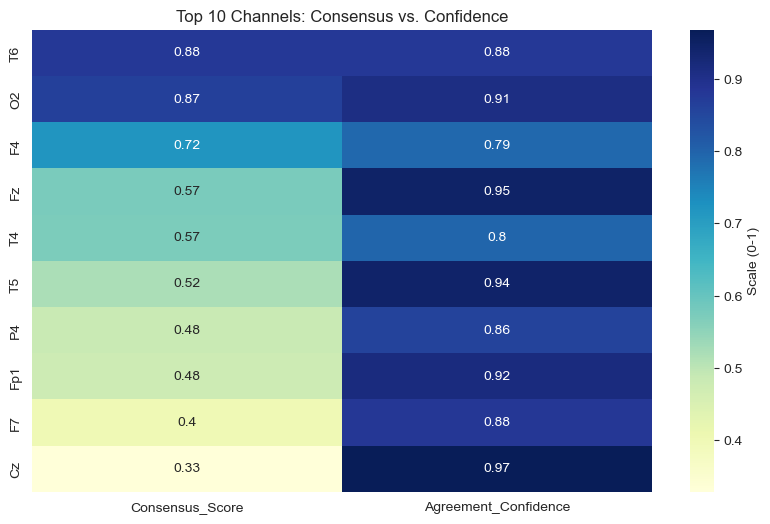}
    \caption{Final consensus and agreement-confidence profile highlighting the top 10 most influential EEG channels aggregated across all evaluated XAI methods.}
    \label{fig:consensus}
\end{figure}

To summarize the attribution patterns identified across the evaluated explainability methods, a final aggregation procedure was performed to derive a consensus channel-ranking profile. The resulting top-ranked channels, together with their corresponding consensus and agreement-confidence scores, are presented in Figure~\ref{fig:consensus}.

The aggregated rankings consistently prioritized the right temporal-parietal channel T6 and the right occipital channel O2, followed by the right frontal channel F4, the midline frontal channel Fz, and the right temporal channel T4. These channels also exhibited comparatively high agreement-confidence scores across the evaluated explainability methods.

Overall, the consensus analysis indicates that several frontal, temporal, and posterior channels were repeatedly emphasized across multiple attribution frameworks despite methodological differences between individual explainability approaches. At the same time, the variability observed between methods suggests that the resulting attribution patterns should be interpreted as method-dependent explanatory estimates rather than definitive neurophysiological biomarkers.

\section{Discussion}
The explainability analysis revealed recurring attribution patterns across multiple post-hoc methods, with consistent emphasis on frontal, temporal, and posterior EEG regions. In particular, several explainability approaches repeatedly prioritized the right frontal ($F_4$, $F_z$), right temporal-parietal ($T_6$), and right occipital ($O_2$) channels, although the relative attribution strengths and ranking order varied between methods. These spatial attribution trends were accompanied by elevated relevance across slower delta-, theta-, and alpha-band activity within frontal regions, as well as prominent beta-band contributions across frontal and temporal regions.

Although post-hoc attribution maps should not be interpreted as direct neurophysiological evidence, the observed patterns are broadly consistent with findings previously reported in EEG studies of Major Depressive Disorder (MDD). Prior resting-state EEG research has frequently associated frontal asymmetry and altered right frontal activity with disrupted emotional regulation and cognitive control processes in depression \citep{Davidson1998,Allen2018MDD,DeLaTorreLuque2017EEGMDD}. The right-lateralized frontal emphasis observed in the present attribution profiles is also compatible with reports of altered hemispheric organization and frontal network changes in depressive populations \citep{DU2023254,TENG202280}. Similarly, the repeated attribution emphasis on posterior regions, particularly the right temporal-parietal ($T_6$) and right occipital ($O_2$) channels, is consistent with reported alterations in posterior alpha- and beta-band activity \citep{FINGELKURTS2006133,Cukic2020EEGMDD,Simmatis2023EEGbiomarkersMDD}. The attribution patterns observed in the present study therefore appear compatible with several previously reported EEG-related findings in MDD, although the current analysis does not establish causal neurophysiological relationships.

A central observation of the study is that different explainability methods produced partially convergent, but not identical, attribution profiles. Gradient-based methods such as Integrated Gradients and GradCAM demonstrated particularly strong agreement, while perturbation-based approaches also exhibited substantial similarity in feature rankings. In contrast, DeepSHAP produced comparatively distinct attribution distributions. These differences likely reflect the fundamentally different assumptions underlying the evaluated explainability frameworks. Gradient-based approaches estimate local sensitivity with respect to the input signal and may be affected by gradient saturation effects in regions where activation responses become relatively flat \citep{sundararajan2017axiomatic}. In contrast, perturbation-based methods such as Occlusion and Permutation evaluate feature importance through controlled input modification, which may alter the intrinsic spatio-temporal structure of EEG recordings and introduce out-of-distribution perturbations \citep{hooker2019benchmark}. Shapley-based approaches further differ by estimating feature contributions through cooperative interaction modeling across feature subsets. Consequently, the observed variability across methods highlights that post-hoc explanations are influenced not only by the learned model representations, but also by the assumptions and operational characteristics of the explainability framework itself.

At the same time, the cross-method agreement observed across several highly ranked channels suggests that some attribution patterns remained relatively stable despite methodological differences. The consensus analysis consistently emphasized right frontal and posterior regions across multiple explainability approaches, while fold-wise variability metrics remained within a relatively narrow range under the adopted evaluation framework. However, these findings should be interpreted cautiously. Stable attribution rankings do not necessarily imply that the identified regions represent definitive biomarkers of depression, nor do they guarantee that the explainability methods fully capture the internal reasoning processes of the deep learning model.

Several additional limitations should also be considered when interpreting the presented results. First, the study relies on a single dataset with a relatively limited cohort size, which restricts generalizability and increases sensitivity to dataset-specific characteristics. Second, although subject-level cross-validation was used to reduce information leakage, EEG recordings remain highly variable and sensitive to acquisition conditions, preprocessing procedures, and inter-subject variability. Third, the analysis focused primarily on aggregated global attribution profiles, while subject-specific explanations and transient temporal relevance patterns were not investigated in detail. Finally, post-hoc explainability methods themselves remain imperfect approximations of model behavior and may produce unstable or partially contradictory explanations depending on baseline selection, perturbation strategy, or attribution formulation.

Future work should therefore focus on validation across larger and more heterogeneous multi-site EEG datasets, together with more systematic investigation of explanation stability under different preprocessing pipelines and attribution configurations. Additional research may also explore inherently interpretable architectures, connectivity-aware transformer models, or approaches incorporating neurophysiological priors directly into the learning process. Such extensions could help clarify which observed attribution patterns reflect reproducible neurophysiological structure and which arise primarily from methodological characteristics of the explainability framework itself.

\section{Conclusion}

This study investigated the behavior of multiple post-hoc explainability methods applied to an InceptionTime model for EEG-based Major Depressive Disorder (MDD) classification. The analysis included Shapley-based, gradient-based, and perturbation-based attribution approaches, enabling comparison of explanation patterns generated under different methodological assumptions.

Across the evaluated explainability frameworks, several recurring attribution trends were observed, particularly involving right frontal, temporal, and posterior EEG regions. Although the relative attribution strengths differed between methods, multiple approaches demonstrated partially convergent channel-ranking profiles and moderate agreement across folds. At the same time, the results also highlighted substantial method-dependent variability, especially between gradient-based and Shapley-based explanations, emphasizing that post-hoc attribution outputs remain strongly influenced by the assumptions and operational characteristics of the selected explainability framework.

The observed attribution patterns were broadly consistent with several findings previously reported in EEG studies of depression, particularly regarding altered frontal asymmetry and posterior rhythmic activity. However, the present results should be interpreted cautiously. The study does not establish causal neurophysiological biomarkers, nor does it demonstrate clinical validity of the generated explanations. Instead, the findings primarily indicate that different explainability methods may capture partially overlapping relevance structures when applied to EEG-based deep learning models for depression detection.

Overall, the study demonstrates the potential usefulness of comparative explainability analysis for investigating the behavior of black-box EEG classifiers, while also illustrating important limitations and sources of uncertainty associated with current post-hoc explainability techniques. Future work should therefore focus on larger multi-site datasets, additional validation strategies, subject-level explanation analysis, and development of more robust and inherently interpretable architectures for biomedical time-series classification.

\bibliographystyle{unsrtnat}
\bibliography{xaireferences}

\end{document}